\documentclass{llncs}
\usepackage[utf8]{inputenc}
\usepackage{mymacros}
\usepackage[textwidth=2.1cm]{todonotes} 
\usepackage{booktabs}
\usepackage{enumitem}
\usepackage{CJKutf8}

\newcommand{\pmblang}{$L_\textnormal{\scriptsize\texttt{PMB}}$}

\title{The Parallel Meaning Bank:\\
A Framework for Semantically Annotating Multiple Languages}

\author{Lasha Abzianidze \and Rik van Noord \and Chunliu Wang \and Johan Bos}

\institute{Center for Language and Cognition, University of Groningen, the Netherlands\\
\email{\{l.abzianidze, r.i.k.van.noord, chunliu.wang, johan.bos\}@rug.nl}
}

\begin{document}

\maketitle

\begin{abstract} 
This paper gives a general description of the ideas behind the Parallel Meaning Bank,
a framework with the aim to provide an easy way to annotate compositional semantics for texts written in languages other than English. The annotation procedure is semi-automatic, and comprises seven layers of linguistic information: segmentation, symbolisation, semantic tagging, word sense disambiguation, syntactic structure, thematic role labelling, and co-reference. New languages can be added to the meaning bank as long as the documents are based on translations from English, but also introduce new interesting challenges on the linguistics assumptions underlying the Parallel Meaning Bank.

\end{abstract}

\section{Introduction}

The Parallel Meaning Bank (PMB) is a semantically annotated parallel
corpus for English, Dutch, German, Italian, Chinese, and Japanese. The key idea behind the PMB is based on the assumption that translations---at least to a large extent---preserve the meaning between the source and target language. Making use of translated texts, annotation for one language can be re-used for the translations, resulting in an economical annotation platform. 
One of the core ideas is that the human annotations can help improve existing language technology (based on supervised machine learning) in the areas of machine translation, automatic question answering and advanced information retrieval.

\begin{figure}[t]
    \centering
    \renewcommand{\arraystretch}{2}
    \begin{tabular}[b]{r @{~~~} l}
    \texttt{EN} & Alfred Nobel invented dynamite in 1866.\\
    \texttt{DE} & Alfred Nobel erfand 1866 das Dynamit.\\
    \texttt{IT} & Alfred Nobel inventò la dinamite nel 1866.\\
    \texttt{NL} & Alfred Nobel vond in 1866 het dynamiet uit.
    \end{tabular} ~
    \raisebox{-9mm}{
    \includegraphics[scale=.9]{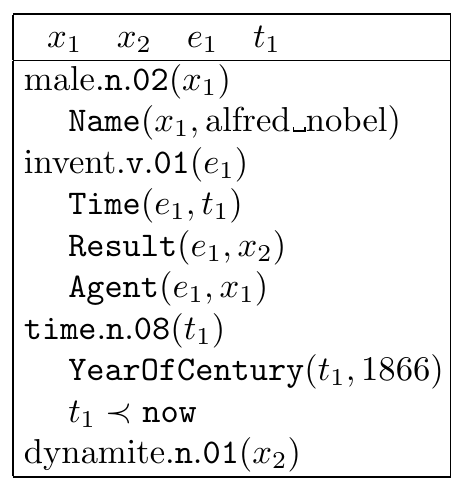}
    }
    \caption{\href{https://pmb.let.rug.nl/explorer/explore.php?part=03&doc_id=0766&type=der.xml}{03/0766}
    PMB document has four meaning-preserving translations. As a result, each translation is annotated with the same meaning representation. 
    }
    \label{fig:drs_03_0766}
\end{figure}

The PMB can be viewed as a 
multilingual version of the Groningen Meaning Bank, GMB
\cite{gmb:eacl,GMB:2017}, an annotation platform designed for the meaning of English texts. Like the GMB, the PMB
contains the raw texts and various layers of linguistic annotation, ultimately resulting in a formal meaning representation based on Discourse Representation Theory (DRT) \cite{kampreyle:drt}.
The annotations are automatically generated by a pipeline of state-of-the-art
natural language processing (NLP) tools and then manually corrected by annotators.
Semantic annotation is hard, even for trained linguists. To give an idea what a meaning representation in the PMB looks like, consider Figure~\ref{fig:drs_03_0766}.
These representations are called Discourse Representation Structures (DRSs) in DRT.

This paper gives a general overview of the PMB and describes several aspects of it in more details.
First, we describe the seven annotation layers that are used to automatically obtain formal meaning representations (Section~\ref{sec:layers} and Section~\ref{sec:pipeline}).
Then, we sketch how the semantic annotation can be projected from one language to another (Section~\ref{sec:projection}).
This is followed by an overview of applications of the released PMB data (Section~\ref{sec:applications}).
Finally, we show how new documents in new languages are added to the PMB and how language technology tools are bootstrapped for new languages (Section~\ref{sec:adding}). 

\section{The Seven Annotation Layers}
\label{sec:layers}

There are two main approaches on semantic annotation. The first approach is to go directly from the source text to the target meaning representations, without any layer of analysis in between. An example of this method is the corpus constructed for Abstract Meaning Representations \cite{amr:13}. The second approach, adopted in the PMB, is to view annotation as a sequence of layers of analysis, where each layer builds on the previous layer by adding a piece of (semantic) information to it. In the PMB, seven layers of annotation are distinguished: 

\begin{enumerate}
\item Tokenisation: detecting sentence boundaries and word tokens;
\item Symbolisation: assigning a non-logical symbol to a word (or multi-word) token. This layer unifies lemmatization and normalization.
\item Word sense disambiguation: assigning concepts to symbols, based on the WordNet \cite{wordnet} sense inventory;
\item Co-reference resolution: marking antecedents for anaphoric expressions;
\item Thematic role labelling: annotate relations between entities using VerbNet roles \cite{Bonial:11} and comparison operators (e.g., temporal and spatial orders);
\item Syntactic analysis: providing lexical categories for each token and building a syntactic structure for the sentence, based on Combinatory Categorial Grammar \cite{Steedman:01};
\item Semantic tagging: assigning a semantic type to a word token \cite{semantic-tagset:17}.
\end{enumerate}

\begin{figure}[t]
\includegraphics[width=\textwidth]{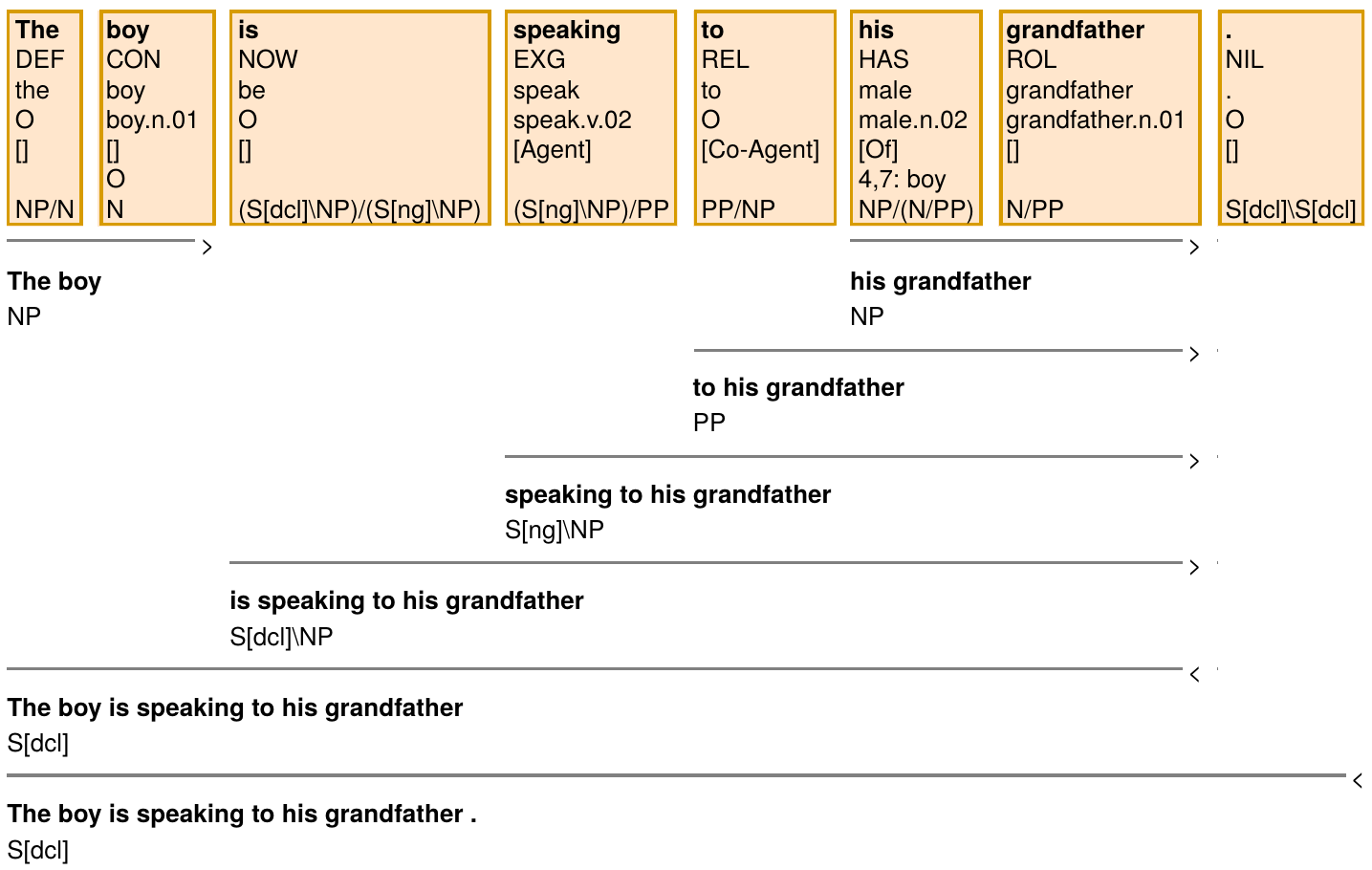}
\caption{All the seven annotation layers of the English translation of
    \href{https://pmb.let.rug.nl/explorer/explore.php?part=03&doc_id=0766&type=der.xml}{46/2924}
    PMB document. The order of layers starting from top:
    tokenisation, semantic tagging, symbolisation, word sense disambiguation, thematic role labelling, co-reference resolution, and syntactic analysis.
    }
\label{fig:layers_46_2924}
\end{figure}

These annotation layers are demonstrated in Figure~\ref{fig:layers_46_2924}.
The annotation layers provide all information needed to provide a compositional semantic analysis for a sentence (for additional details about the PMB annotation layers see \cite{PMBshort:2017}).
This is done by using the lambda calculus, and adopting Discourse Representation Theory as semantic formalism, implemented by the semantic parser Boxer \cite{boxer}. In a final step, the semantic analysis of single sentences are combined into one meaning representation covering the entire text.

\section{Annotation Pipeline}\label{sec:pipeline}

\begin{figure}[t]
    \includegraphics[width=\textwidth]{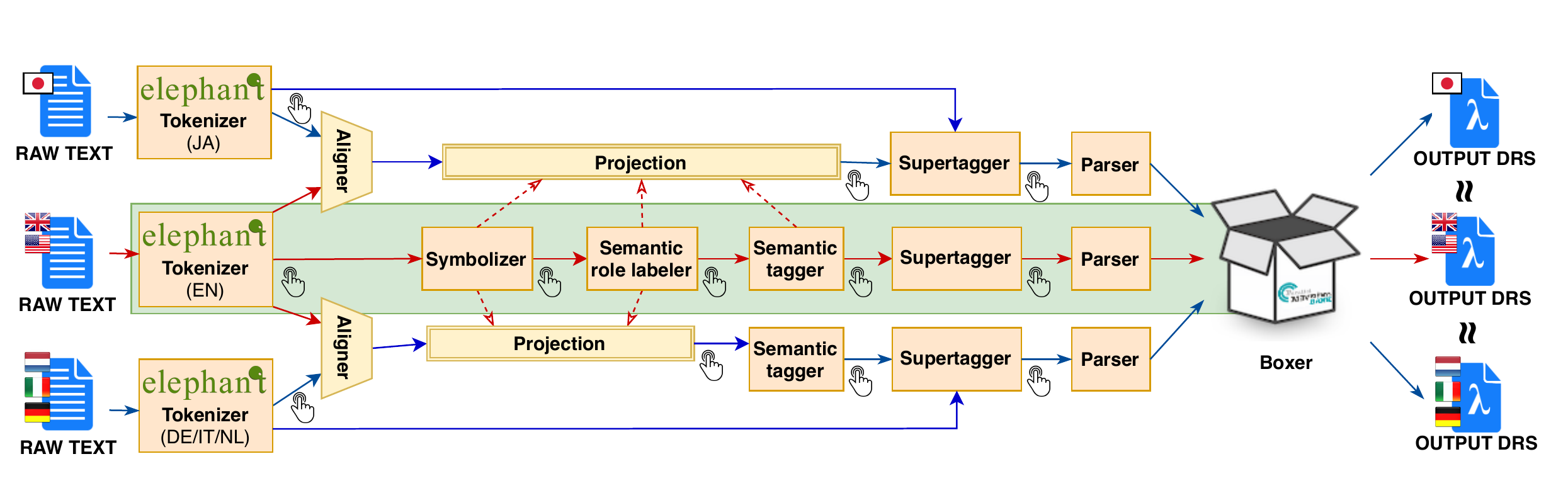}
    \caption{The PMB pipeline: a sequence of NLP tools that processes raw texts and outputs formal meaning representations.
    The hand icon indicates functionality of overwriting parts of the system outputs with manual annotations.}
    \label{fig:pipeline}
\end{figure}

Manually creating the seven annotation layers for a large amount of documents is not a feasible task.
For this reason, we use an annotation pipeline to automatically segment raw documents, label tokens with token-based annotations, and produce the final meaning representation.
The pipeline consists of a sequence of NLP tools each serving for a specific annotation layer. 
The pipeline of English-specific tools is highlighted with a green background in Figure~\ref{fig:pipeline}.
Below, we briefly describe each NLP tool:%
\footnote{Currently, the pipeline lacks specialized NLP tools for word sense disambiguation and co-reference resolution.
Therefore, these layers are manually annotated for now.}

\begin{itemize}
\item Elephant \cite{elephant} is used for tokenisation. The tool performs sentence boundary and word token detection as a single labelling task: each character is labelled with one of the four labels depending on being sentence beginning, token beginning, inside token, and outside token;
\item Semantic tagging is carried using the tri-gram based TnT tagger \cite{Brants:2000};
\item The lemmatisation part of symbolisation is done with the help of the lemmatizer Morpha \cite{minnen:01}.
Currently, we use instance-based learning for the normalisation part.
In particular, for every existing combination of lemma and semantic tag in the PMB, the most frequent symbol is memorized which is later reused to tag a token with the corresponding pair of semantic tag and lemma.
For example, to get a symbol for a token \textit{eight}, first its lemma \texttt{eight} and semantic tags \texttt{QUC} is obtained and then the instance-based learning will assign \texttt{8} as a symbol to it
\item Obtaining syntactic analysis consists of assigning lexical categories to tokens and constructing a derivation tree over these categories.
The both subtasks are performed using EasyCCG \cite{lewisSteedman:14}, a CCG-based parser that requires only tokenised input and pre-trained word embeddings.
\item Thematic role labelling is done with a tagger based on Conditional Random Fields \cite{crf:01}. The tagger employs semantic tags, symbols and CCG lexical categories as features to predict thematic roles.
\end{itemize}

The output of each tool can be manually corrected by human annotators.%
\footnote{The PMB documents can be manually annotated with the PMB explorer, an online annotation environment, available at: \url{https://pmb.let.rug.nl/explorer}. Anybody can register and annotate the documents.}
In this way, we use a human-in-the-loop approach 
to obtain gold standard annotation layers and the final meaning representations.
We also apply bootstrapping with the gold standard annotation layers to retrain and further improve the quality of the NLP tools.
This aims at reducing human annotation efforts while still retaining high quality system outputs.

\section{Annotation Projection}\label{sec:projection}

The previous two sections gave a rough overview of what is required to provide a compositional analysis for the meaning of a text for one language. For historical reasons, this language is English, because of the tools developed earlier in the Groningen Meaning Bank \cite{gmb:lrec}. Instead of starting from scratch and implementing a pipeline for other languages, we follow a different approach in the PMB. This approach is called \textit{annotation projection}, and requires that the English text has an adequate translation in the language of your choice. The first languages that we added in the PMB were languages close to English, such as other Germanic languages (Dutch and German) and Italian, a Romance language. 

\begin{figure}[t]
\includegraphics[scale=.84]{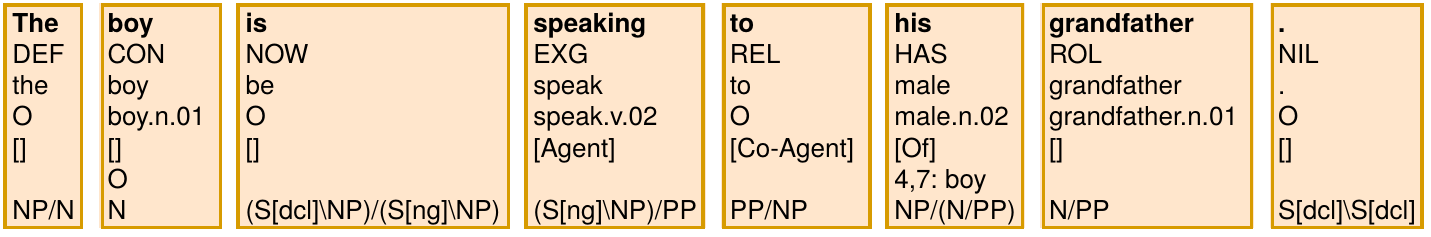}\\[-1.1mm]
\hspace*{3.5mm}\includegraphics[scale=.557]{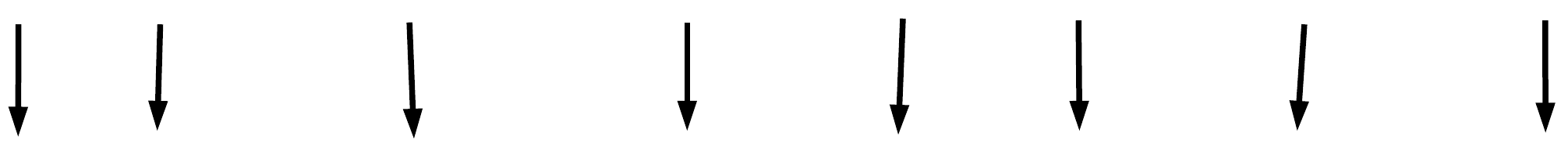}\\[-2mm]
\includegraphics[width=\textwidth]{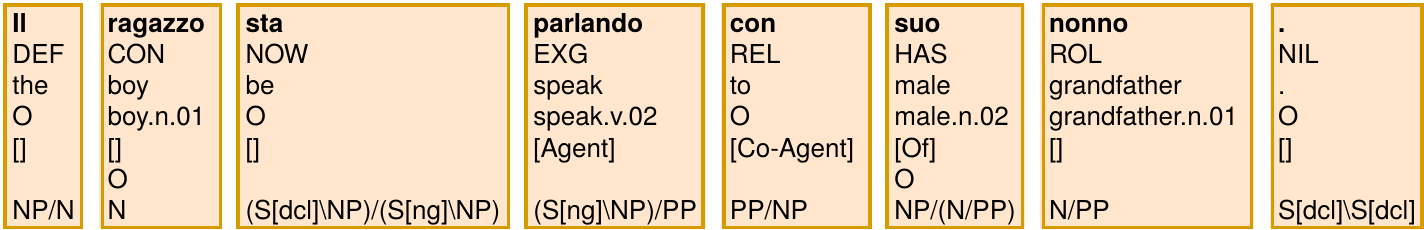}
\caption{An example of a complete annotation projection: all the seven annotation layers are projected from English to Italian.}
\label{fig:projection_boy}
\end{figure}

The idea of semantic projection is extremely simple, but its implementation is surprisingly challenging even for closely-related languages. The assumption that a translation doesn't change much of the meaning, is the driving force in this approach. But for reasons of scalability, we are not just interested in the final meaning representation, but also in the compositional analysis supporting this final meaning representation. This makes projection more challenging.

In the PMB, annotation projection is implemented using word alignment between English and the target language.%
\footnote{We employ GIZA++ \cite{giza} to automatically induce word alignments.}
The alignments provide clues how to transfer the layers of annotation from English to the other languages \cite{evang-thesis}. For cases where the syntactic structure of the target language is similar to that of the source language (English), this is often straightforward. Figure~\ref{fig:projection_boy} shows one of such cases where a literal translation leads to a perfect word alignment and therefore to a complete annotation projection.
This leads to the very same meaning representation for the Italian translation that the English translation had.

But translations are not always in a perfect word-to-word and order-preserving correspondence as in the previous example. Even closely-related language show different behaviour with respect to a word order, multi-word expressions, definiteness, use of articles, and noun-noun compounds.
So automatic projection requires the help of human annotators to provide corrections.

\begin{figure}[t]
\includegraphics[scale=.773]{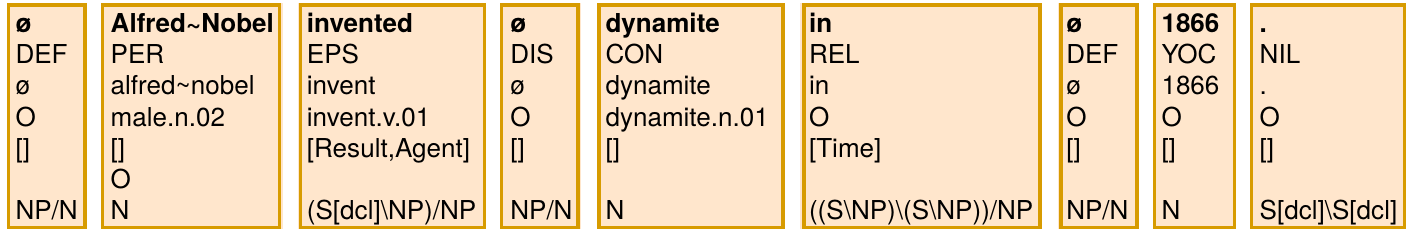}\\[-1mm]
\hspace*{14mm}\includegraphics[scale=.51]{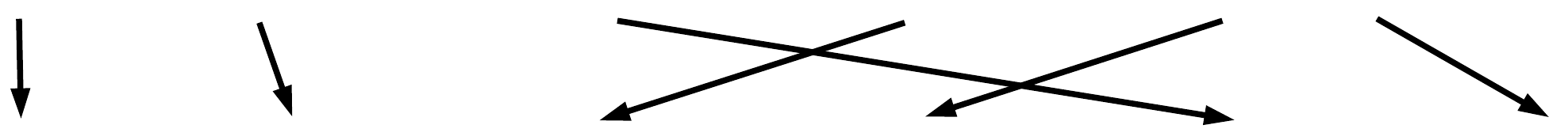}\\[-1mm]
\includegraphics[width=\textwidth]{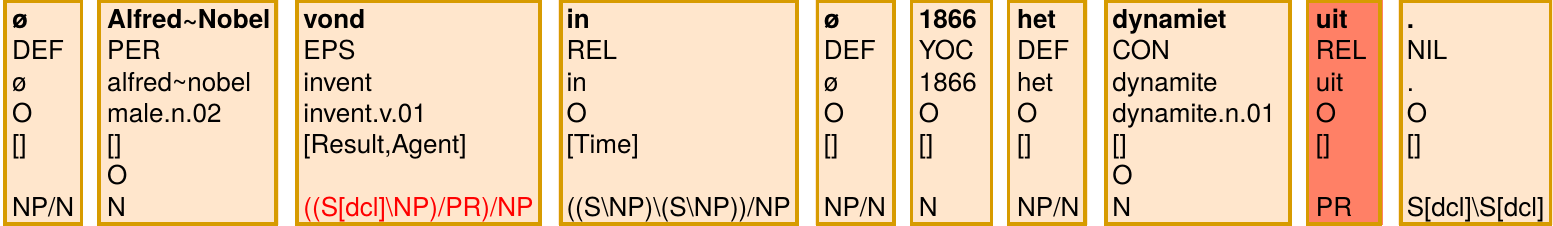}
\caption{An imperfect annotation projection is compensated by the language-specific syntactic parsing model.}
\label{fig:projection_alfred}
\end{figure}

In the PMB, we go beyond the mere annotation projection as it is brittle for wide-coverage translations.
To do so, using the same NLP tools, we (re-)train semantic tagging and syntactic parsing models for non-English languages.
Initially the training data consisted of translations with perfect annotation projections.
Gradually the training data increased as a result of reprocessing the rest of the translations with new models and correcting manually where necessary.
For example, the annotation projection in Figure~\ref{fig:projection_alfred} fails for the syntactic analysis layer due to the difference in a word order of the Dutch translation.
But with the help of the in-house trained Dutch model of the parser, it is possible to automatically recover a correct syntactic analysis of the Dutch translation, which eventually leads to the same meaning representation (see Figure~\ref{fig:drs_03_0766}).%
\footnote{To verify whether projected annotations yield the same meaning representation as of English, we perform fine-grained matching of meaning representations \cite{lrec:2018}.
} 

Figure~\ref{sec:pipeline} shows the PMB pipeline of NLP tools that simultaneously processes documents in five languages.
While currently only symbols and thematic roles are projected for the Dutch, German, and Italian translations, the Japanese translations also get semantic tags projected from the English translations.
In the near future, we plan to retrain Japanese-specific model for the semantic tagging.

Currently we are investigating what consequences semantic annotation projection  has on languages that behave significantly different from English from a linguistic perspective. Here we think of languages such as Chinese and Japanese, and perhaps also Kartvelian languages such as Georgian \cite{GeoCorpus:14}. These languages add pressure on the principles of the PMB, in particular on the extent one can adopt a single framework for each layer in the semantic analysis pipeline. 
To give a first example, the semantic tags might be subject to extension of the tagset for new languages that show phenomena that cannot be captured with the existing semantic categories. 
To give a second example, we assume CCG as the theory of syntatic structure suitable for all languages. CCG starts with a base of atomic categories, which work well for Germanic languages, but other languages could be hard to adopt in the parameters provided for English. In future work we need to take a closer look at such a wider perspective.
As a final example, in Chinese, there are less syntactic constraints for verbs, but there is  widespread use of pro-drop, and a larger distribution of ambiguous constructions, such as the relative clause and verbal coordination. In addition, the inherent ambiguities caused by both verbal coordination and relative clauses of Chinese make semantic parsing more difficult than syntactic parsing \cite{yu-etal-2011-analysis}.

\section{Applications}\label{sec:applications}

The PMB annotations are released periodically, free of charge.\footnote{\url{https://pmb.let.rug.nl/data.php}} It includes gold standard data, which is fully manually corrected, as well as silver (partially manually corrected) and bronze (with no manual corrections) data. The releases so far contain documents for English, German, Italian and Dutch, but for future releases we plan to include Chinese and Japanese. An overview of the releases is shown in Table~\ref{tab:releases}.

\begin{table}[!htb]
\caption{Number of released documents per language for the five current PMB releases.\label{tab:releases}}
\centering
\begin{tabular}[t]{llr@{\kern7mm}r@{\kern7mm}r@{\kern7mm}r}
\toprule
\textbf{Release} \hspace*{10mm} & \textbf{Quality} \hspace*{10mm} & \textbf{EN} & \textbf{DE} & \textbf{IT} & \textbf{NL} \\
\midrule                
\textbf{PMB-1.0.0} & \textbf{Gold}   &  2,049           & 641             &  387           & 394            \\
\midrule
\textbf{PMB-2.0.0} & \textbf{Gold}   &  3,925           & 1,048            &     568        &  527           \\
 & \textbf{Silver} &   66,693          &   611          &      266       &      192       \\
\midrule

\textbf{PMB-2.1.0} &  \textbf{Gold}   &   4,555          & 1,175             &    635         &     586        \\
& \textbf{Silver} &   71,308          &    688         &   306          &      207       \\
\midrule
\textbf{PMB-2.2.0} & \textbf{Gold}   &     5,929        &   1,419          &  724          &       633      \\
& \textbf{Silver} &   67,965          &   4,235          &   2,515          &  1,051           \\
& \textbf{Bronze} &  120,622           &   102,998          &     61,504        &  20,554     \\

\midrule                

\textbf{PMB-3.0.0} & \textbf{Gold}   &  8,403           &  1,979           &     1,062        &    1,012         \\
& \textbf{Silver} &  97.598           &    5,250         &    2,772         &     1,301        \\
& \textbf{Bronze} &   146,371          &    121,111         &    64,305         & 21,550      \\
\bottomrule
\end{tabular}
\end{table}

One of the goals of the PMB releases is to aid DRS parsing, a task in which a model has to automatically produce a DRS from raw text. These produced DRSs can then potentially be of benefit in other language related tasks, such as machine translation or question answering. Early approaches used rule-based system for only small fragments of English \cite{johnsonklein,wadaasher}, though wide-coverage semantic parsers that use supervised machine learning were also developed, mainly on the GMB data \cite{step2008:boxer,le:12,boxer,neural_drs_gmb:18,liu-etal-2019-discourse-representation}.

The main advantage of the PMB is that it contains gold standard data for evaluating the parsers. This is in contrast to the GMB, which contains partially manually corrected evaluation sets that are not guaranteed to be gold standard. This allowed for the organization of a shared task on English DRS parsing in PMB format \cite{abzianidze-etal-2019-first}. Five systems participated in this shared task, which all used neural networks in some capacity. Three systems used sequence-to-sequence models based on the first PMB-based DRS parser \cite{van-noord-etal-2018-exploring}, which was extended by including linguistic features \cite{drsiwcs:19,W19:noord} and by swapping the bi-LSTM encoder/decoder for a transformer model \cite{W19:liu}, which was the winning system. The two other systems consisted of a transition-based parser that relied on stack-LSTMs \cite{W19:evang} and a neural graph parsing system that converted the DRSs to a more general graph format before parsing \cite{fancellu-etal-2019-semantic}. The latter is also the first system that produced results for German, Italian and Dutch DRS parsing.

There are also other applications of the PMB data. For one, semantic tagging can be useful as either an auxiliary task to improve a main task \cite{Bjervaetal:16,bjerva2017will,abdou2018can}, or as a general dataset for evaluating neural architectures \cite{belinkov2017evaluating,liu-etal-2019-linguistic,dalvi2020exploiting,ek2019language}. Moreover, PMB data has been used in research on natural language inference \cite{yanaka2019can} and machine translation \cite{durrani-etal-2019-one}.

\section{A Look at the Future: Extending the PMB}\label{sec:adding}

The PMB can be extended in terms of introducing new documents or new translations.
Translations may belong to languages that are new or already covered in the PMB.
In case a translation belongs to a new language, its integration in the PMB requires more work as the new language needs to be processed by the PMB pipeline.
In this section, we describe the procedure and conditions for extending the PMB.

The simplest extension procedure is when adding translations to PMB documents in one of the PMB (non-English) languages, let's say \pmblang{}.
In this case, no new documents are created, and there is no need to develop new NLP tools as the PMB pipeline can already process texts in \pmblang{}.
If the PMB uses the projection method for \pmblang{}, then it is necessary to align the new \pmblang{}-translations to the existing English translations.
For the best results, the alignment is usually done on all PMB English-\pmblang{} bitexts.
This might affect the alignments of old PMB documents and annotations of the projected layers, consequently.
Since the alignment is carried out on more bitexts than before, the assumption is that the quality of alignments improves.
Whether the change influences alignments negatively, this can be verified for the translations already having a gold standard annotation for the projected layers.
The difference for the projected layers will show up as conflicts with the gold standard.

Adding a new parallel corpus to the PMB involves adding completely new documents.
Taking the architecture of the PMB into account, one of the languages of the new corpus must be English.
Let's first consider the scenario when all the languages of the corpus are covered by the PMB.
All new documents (consisting of translations) get new \texttt{part/doc} identifiers and are uniformly distributed over all the 100 parts of the PMB. 
If the newly added documents belong to a text genre new to the PMB, some NLP tools in the pipeline might require further adaptation.
For example, if the documents belong to the social media domain, one might need to correct the tokenization or semantic tagging of slang words and retrain the corresponding tools on the corrected annotations.
Additionally, the procedures of inducing new alignments and verifying the changes caused by them are also applicable in this scenario.

The case where newly added parallel corpus contains translations not belonging to the PMB languages is the most laborious.
New languages require their own annotation pipelines.
Here, we describe our first experiences from adding Japanese \cite{yanaka2020nlp}
and Chinese, using  translations from Tatoeba.\footnote{\url{https://tatoeba.org}}

To enable the annotation projection from English to Japanese, it is necessary to extract word alignments from the bitext, which itself presupposes tokenisation of the Japanese translations.
Since we strive to use the same NLP tools with language-specific models for each annotation layer, we trained a Japanese model of the Elephant tokenizer.\footnote{The training data was obtained by processing the Japanese translations in the PMB with the UDPipe 1.2.0 \cite{straka-etal-2016-udpipe} and the model \texttt{japanese-gsd-ud-2.3-181115}.}
After extracting the word alignments, token-based annotations were projected for one-to-one word alignments.
Since English and Japanese are languages with radically different typologies, the annotation projection for the syntactic analysis failed for almost all Japanese translations.
As
syntactic analyses play a key role for obtaining meaning representations in the PMB because they contribute to defining lexical semantics and guiding composition of phrasal semantics,
 a quick integration required a Japanese CCG parser in the PMB pipeline.
Fortunately, there exists a Japanese CCG parser, depCCG \cite{yoshikawa-etal-2017-ccg}.
We trained a new Japanese model for EasyCCG on the output of depCCG.
We opted for training a new model to keep the PMB pipeline lean rather than integrating an additional tool in it. 
In the near future, we plan to train a Japanese model for the semantic tagging in order to eliminate ``holes'' in the semantic tagging layer caused by the annotation projection.

We are currently adding (Mandarin) Chinese translations for the PMB documents.
While doing so, we are taking a route similar to the one we took for Japanese.
To train the Chinese model for Elephant, we used the output from jieba.\footnote{\url{https://github.com/fxsjy/jieba}} 
The EasyCCG model was trained on the CCG derivation trees which were obtained from the Chinese Treebank \cite{ChTB} following \cite{tse-curran-2010-chinese}.

\medskip

The current undertakings of adding more languages to the framework doesn't mean that all problems are solved. 
The entire PMB enterprise emits a formal flavour of universality of language analysis. 
This is reflected in the practical use of our language technology pipeline, with the aim of using the same NLP tools but employing the language-specific models as the only variable element. We have reached a high level of generalization, but there are also many refinements that seek improvement, in particular on the ontological, categorial, and contextual level. The only way to make progress in this area of computational semantics is by considering other languages and getting your hands dirty!

\bibliographystyle{splncs03.bst} 
\bibliography{mybib}

\end{document}